\PassOptionsToPackage{unicode}{hyperref}
\PassOptionsToPackage{hyphens}{url}
\documentclass[
  letterpaper]{article}
\usepackage{xcolor}
\usepackage{amsmath,amssymb}
\usepackage{iftex}
\ifPDFTeX
  \usepackage[T1]{fontenc}
  \usepackage[utf8]{inputenc}
  \usepackage{textcomp} 
\else 
  \usepackage{unicode-math} 
  \defaultfontfeatures{Scale=MatchLowercase}
  \defaultfontfeatures[\rmfamily]{Ligatures=TeX,Scale=1}
\fi
\usepackage{lmodern}
\ifPDFTeX\else
\fi
\IfFileExists{upquote.sty}{\usepackage{upquote}}{}
\IfFileExists{microtype.sty}{
  \usepackage[]{microtype}
  \UseMicrotypeSet[protrusion]{basicmath} 
}{}
\makeatletter
\@ifundefined{KOMAClassName}{
  \IfFileExists{parskip.sty}{%
    \usepackage{parskip}
  }{
    \setlength{\parindent}{0pt}
    \setlength{\parskip}{6pt plus 2pt minus 1pt}}
}{
  \KOMAoptions{parskip=half}}
\makeatother
\usepackage{color}
\usepackage{fancyvrb}

\DefineVerbatimEnvironment{Highlighting}{Verbatim}{commandchars=\\\{\}}
\newenvironment{Shaded}{}{}

\newcommand{\NormalTok}[1]{#1}

\usepackage{graphicx}
\makeatletter
\newsavebox\pandoc@box
\newcommand*\pandocbounded[1]{
  \sbox\pandoc@box{#1}%
  \Gscale@div\@tempa{\textheight}{\dimexpr\ht\pandoc@box+\dp\pandoc@box\relax}%
  \Gscale@div\@tempb{\linewidth}{\wd\pandoc@box}%
  \ifdim\@tempb\p@<\@tempa\p@\let\@tempa\@tempb\fi
  \ifdim\@tempa\p@<\p@\scalebox{\@tempa}{\usebox\pandoc@box}%
  \else\usebox{\pandoc@box}%
  \fi%
}
\def\fps@figure{htbp}
\makeatother
\NewDocumentCommand\citeproctext{}{}

\makeatletter
 \let\@cite@ofmt\@firstofone
 \def\@biblabel#1{}
 \def\@cite#1#2{{#1\if@tempswa , #2\fi}}
\makeatother
\newlength{\cslhangindent}
\newlength{\csllabelwidth}
\newenvironment{CSLReferences}[2] 
 {\begin{list}{}{%
  \setlength{\itemindent}{0pt}
  \setlength{\leftmargin}{0pt}
  \setlength{\parsep}{0pt}
  \ifodd #1
   \setlength{\leftmargin}{\cslhangindent}
   \setlength{\itemindent}{-1\cslhangindent}
  \fi
  \setlength{\itemsep}{#2\baselineskip}}}
 {\end{list}}
\usepackage{calc}

\providecommand{\tightlist}{%
  \setlength{\itemsep}{0pt}\setlength{\parskip}{0pt}}
\usepackage[letterpaper,margin=1in]{geometry}
\usepackage{amsmath}
\usepackage{amssymb}
\usepackage{bookmark}
\IfFileExists{xurl.sty}{\usepackage{xurl}}{} 
\hypersetup{
  pdftitle={Distributed Agent Reasoning Across Independent Systems With Strict Data Locality},
  pdfauthor={Daniel Vaughan; Kateřina Vaughan},
  hidelinks,
  pdfcreator={LaTeX via pandoc}}

\title{Distributed Agent Reasoning Across Independent Systems With
Strict Data Locality}
\author{Daniel Vaughan \and Kateřina Vaughan}
\date{20 November 2025}

\begin{document}
\maketitle
\begin{abstract}
This paper presents a proof-of-concept demonstration of agent-to-agent
communication across distributed systems, using only natural-language
messages and without shared identifiers, structured schemas, or
centralised data exchange. The prototype explores how multiple
organisations (represented here as a Clinic, Insurer, and Specialist
Network) can cooperate securely via pseudonymised case tokens, local
data lookups, and controlled operational boundaries. The system uses
Orpius as the underlying platform for multi-agent orchestration, tool
execution, and privacy-preserving communication. All agents communicate
through OperationRelay calls, exchanging concise natural-language
summaries. Each agent operates on its own data (such as synthetic clinic
records, insurance enrolment tables, and clinical guidance extracts),
and none receives or reconstructs patient identity. The Clinic computes
an HMAC-based pseudonymous token, the Insurer evaluates coverage rules
and consults the Specialist agent, and the Specialist returns an
appropriateness recommendation. The goal of this prototype is
intentionally limited: to demonstrate feasibility, not to provide a
clinically validated, production-ready system. No clinician review was
conducted, and no evaluation beyond basic functional runs was performed.
The work highlights architectural patterns, privacy considerations, and
communication flows that enable distributed reasoning among specialised
agents while keeping data local to each organisation. We conclude by
outlining opportunities for more rigorous evaluation and future research
in decentralised multi-agent systems.
\end{abstract}

\section{Introduction}\label{introduction}

Large language models (LLMs) have accelerated research into multi-agent
systems capable of reasoning, collaboration, and task delegation. Most
existing frameworks, however, assume a single system boundary with
shared memory, unified context windows, or globally accessible
identifiers. Real organisations do not operate this way: data cannot be
freely shared, and collaboration must respect privacy, tenancy
boundaries, and heterogeneous infrastructure.

This paper describes a proof-of-concept demonstration in which three
independent organisations (\emph{Clinic}, \emph{Insurer}, and
\emph{Specialist Network}) coordinate using LLM-driven agents deployed
on separate Orpius\footnote{Orpius is a software platform that provides
  the execution environment for the prototype described here, including
  agent configuration, tool execution, and cross-node operation calls.
  Developed by Outcoder Sàrl. Further information: https://orpius.com.}
systems. Although the system forms a distributed multi-agent
environment, it also behaves as a federated system in the operational
sense. Each organisation maintains exclusive control over its own data,
tools, and operational policies, and participates only through
controlled natural-language summaries. Coordination occurs without
shared schemas, shared identifiers, or direct data exchange. This
creates a practical model of federated decision support rather than
federated training, and reflects how real organisations collaborate
without pooling data. This arrangement enables interoperability without
centralisation and preserves organisational autonomy.

The prototype shows that:

\begin{enumerate}
\def\labelenumi{\arabic{enumi}.}
\tightlist
\item
  Agents communicate solely through concise natural-language summaries.
\item
  Sensitive identifiers can be replaced with \emph{HMAC-derived
  pseudonymous tokens}, enabling linkage without disclosure.
\item
  Each organisation enforces strict data locality, with no cross-entity
  access to internal records.
\item
  Orpius's OperationRelay enables structured cross-system communication
  without shared schemas or RPC-style contracts.
\end{enumerate}

The demonstration centres on a deliberately simple yet realistic
scenario: the Clinic assesses a patient and seeks coverage
clarification; the Insurer evaluates the request and, when necessary,
consults a Specialist agent. Throughout, no identifying information or
local dataset is exchanged, and all reasoning occurs on isolated data
held within each organisation's Orpius node.

Importantly, this work does \emph{not} aim to provide medical accuracy,
clinical validation, or a large-scale evaluation. Its contribution lies
in the architecture, message-passing patterns, and privacy-preserving
mechanisms that support distributed natural-language agent interaction.
The synthetic data included serves purely illustrative purposes.

Real organisations rarely operate within a single technical boundary,
and most cannot pool data across departments or partners due to
regulatory, contractual, or operational constraints. Yet many decisions,
especially in healthcare, finance, and public services, depend on
information held by multiple parties. Existing multi-agent frameworks
generally assume a unified environment with shared context or structured
schemas. These assumptions do not hold in inter-organisation settings.
This creates a practical gap: how can autonomous agents collaborate when
participants cannot share data, infrastructure, or execution state? The
prototype explored in this paper demonstrates that meaningful multi-step
reasoning can still occur through controlled natural-language
interactions while maintaining strict data locality.

The system therefore operates as a form of federated decision support.
Independent organisations cooperate by exchanging structured
natural-language requests rather than shared features, model weights, or
structured schemas. All data remain local, no identifiers or structured
records are shared, and the only information that crosses system
boundaries is a concise natural-language summary created by the calling
agent. This supports collaboration under strict data locality while
avoiding the need for any centralised infrastructure.

\pandocbounded{\includegraphics[keepaspectratio,alt={}]{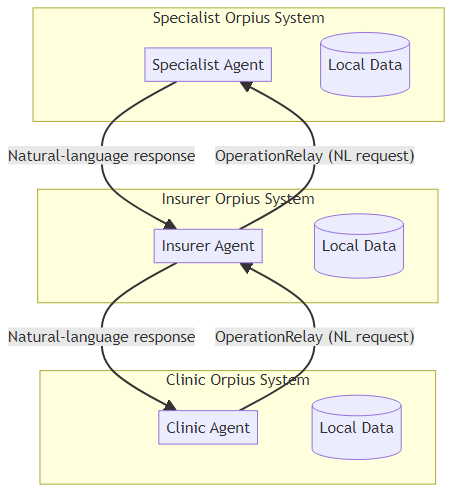}}

\emph{Figure 1. Overview of the prototype architecture. Each
organisation runs its own Orpius system with local agents and local
data. Cross-system cooperation occurs through OperationRelay requests
and natural-language responses. No shared identifiers, schemas, or data
are exchanged.}

\section{Contributions}\label{contributions}

This paper introduces:

\begin{enumerate}
\def\labelenumi{\arabic{enumi}.}
\item
  A novel architecture for distributed agentic reasoning, enabling
  LLM-driven agents on completely independent systems to cooperate
  without shared schemas, shared memory, or centralised infrastructure.
\item
  A privacy-preserving linkage strategy based on HMAC tokens that allows
  cross-organisation coordination without exposing identity.
\item
  An architecture for strict data locality, where each organisation
  retains exclusive access to its own storage, tools, secrets, and
  operational policies.
\item
  A model for \emph{federated decision support}, in which cross-node
  coordination is carried out through controlled OperationRelay calls
  that exchange concise natural-language requests and responses rather
  than structured identifiers or records.
\item
  A practical, end-to-end demonstration showing how these ideas support
  a realistic workflow involving a clinic, an insurer, and a specialist
  network.
\end{enumerate}

\section{Background and Motivation}\label{background-and-motivation}

\subsection{Multi-agent LLM systems and
coordination}\label{multi-agent-llm-systems-and-coordination}

Multi-agent systems built on LLMs are increasingly used for planning,
tool use, delegation, negotiation, and distributed problem solving.
Recent work examines emergent behaviours (Park et al. 2023), task
decomposition ({Du et al.} 2023), and collaborative reasoning (Z. Wang
et al. 2023). Other studies analyse the dynamics of agent communication,
including emergent protocols, iterative optimisation, and negotiation
strategies within shared environments (Liu et al. 2023; L. Wang et al.
2023).

These frameworks typically assume a single technical boundary with
shared memory, global context, or common tool access. Agents may have
different roles, but they operate over a common pool of state. In
contrast, the setting studied in this paper assumes \emph{isolated
organisational nodes} that cannot share identifiers, raw records, or
infrastructure. Agents must still coordinate, but only through
constrained natural-language summaries that respect data locality.

\subsection{Federated and privacy-preserving
computation}\label{federated-and-privacy-preserving-computation}

Federated learning and federated analytics aim to enable joint
computation without centralising raw data (Konečný et al. 2016; McMahan
et al. 2017). In these systems, model parameters or aggregates are
exchanged while data remain on local devices or servers. The core ideas
are data locality, controlled information flow, and limited disclosure.

This work does not implement federated training. Instead, it explores
\emph{federated decision support}: autonomous agents hosted by different
organisations communicate in order to reach a decision, while all
underlying records remain local. The prototype borrows principles from
privacy-preserving computation, such as pseudonymous linkage and data
minimisation (\emph{General Data Protection Regulation} 2016), but
applies them to natural-language workflows rather than numerical
optimisation. A one-way HMAC token links cases across Clinic and Insurer
nodes without revealing identity, and messages contain only the
information needed for the next reasoning step.

\subsection{LLMs in clinical decision
support}\label{llms-in-clinical-decision-support}

LLMs have been proposed for clinical summarisation, triage, and decision
support ({Singhal et al.} 2023; {Nori et al.} 2023). These studies
highlight both potential benefits and safety concerns when models
interact with medical data.

The present prototype does not attempt to extend that body of work with
new clinical findings. It uses \emph{synthetic clinical records} and
\emph{non-validated guideline extracts}, solely to make the architecture
concrete. No claims are made about medical accuracy, safety, or
regulatory suitability. The contribution lies in showing how a
clinical-style workflow can be orchestrated across independent systems
under strict data locality, rather than in evaluating the quality of the
medical advice itself.

\subsection{Agent platforms and execution
environments}\label{agent-platforms-and-execution-environments}

A growing ecosystem of frameworks and platforms (for example Ray,
HuggingFace Autogen, LangChain, and other orchestration libraries)
support agent tool use, parallel execution, and coordination within a
shared environment. These systems often assume that agents can access a
common file system, a shared vector store, or a unified API surface,
even when they are logically separated.

By contrast, many real collaborations span independent organisations
with separate infrastructure and regulatory obligations. Systems cannot
simply share a database or mount a common storage volume. Our prototype
targets this setting by treating each organisation as a separate node
that exposes only a small set of operations for external use. We use
Orpius as the execution environment for each node, because it provides
isolated deployments, a tool-calling loop, and a cross-node
OperationRelay mechanism. The architectural details of this environment
are described in the System Overview and Prototype Architecture
sections; here, the important point is that agents coordinate without a
shared global state or schema, and must rely on natural-language
summaries and pseudonymous tokens instead.

\section{System Overview}\label{system-overview}

The prototype consists of three independent Orpius deployments, each
representing a distinct organisation:

\begin{enumerate}
\def\labelenumi{\arabic{enumi}.}
\tightlist
\item
  \emph{Clinic System}: holds patient observations and proposes a
  treatment.
\item
  \emph{Insurer System}: holds enrolment and coverage rules.
\item
  \emph{Specialist Network System}: holds clinical guidance extracts
  used to assess appropriateness.
\end{enumerate}

\pandocbounded{\includegraphics[keepaspectratio,alt={}]{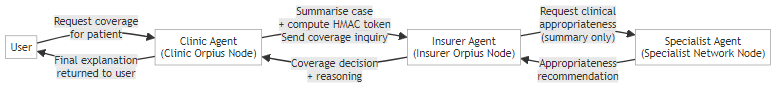}}

\emph{Figure 2: Distributed reasoning loop across Clinic, Insurer, and
Specialist agents. Each agent operates on local data; only
natural-language summaries (and a pseudonymous token between Clinic and
Insurer) cross organisational boundaries.}

\subsection{Orpius as the Execution
Environment}\label{orpius-as-the-execution-environment}

In this prototype, each participating organisation runs its own Orpius
node. Orpius is more than an LLM integration layer; it provides the
execution environment, security model, and orchestration mechanisms that
make cross-node agentic decision making possible while keeping data
local.

The following concepts are sufficient to understand the prototype:

\begin{enumerate}
\def\labelenumi{\arabic{enumi}.}
\item
  \textbf{Node-level isolation and tenancy boundaries} Each Orpius
  deployment is single-tenant and maintains its own compute, storage,
  and secrets. The Clinic, Insurer, and Specialist Network therefore run
  on three separate Orpius nodes, each with isolated storage for CSV
  files, guidance extracts, and configuration. No node can directly read
  the files or databases of another; the only cross-node interaction
  occurs through explicitly configured operations.
\item
  \textbf{Agents, tools, and operations as core abstractions} Orpius
  treats agents as users of the system, with access to a pool of tools.
  Tools include built-in capabilities such as file access, code
  execution, web retrieval, and event handling, and custom tools hosted
  on external services. External systems interact with agents through
  \emph{Operations}. An Operation specifies which agent receives a
  request and which tools that agent may invoke on behalf of the caller.
  In this prototype, each organisation exposes one Operation that fronts
  its local agent (Clinic, Insurer, Specialist). The Operation
  configuration constrains which tools may be used and prevents an agent
  from escaping its intended scope during cross-organisation workflows.
\item
  \textbf{Data locality and secrets management} Each node maintains its
  own isolated file storage. The Clinic Agent reads only its local
  \texttt{patients.csv} and \texttt{clinical\_observations.csv}; the
  Insurer Agent reads only \texttt{enrollment.csv} and
  \texttt{coverage\_rules.csv}; the Specialist Agent accesses only the
  osteoarthritis guidance extracts. Secrets such as the HMAC key used to
  derive the pseudonymous \texttt{patientToken} are stored in a managed
  secret store and referenced symbolically in prompts. The key is
  resolved only at tool-execution time inside the node, never sent to
  the LLM provider or to another node. Local storage, combined with
  secret management, underpins the privacy guarantees described in the
  tokenisation strategy.
\item
  \textbf{Tool-calling orchestration and cross-node OperationRelay}
  Orpius implements a structured loop for model calls: an origin (chat,
  event, schedule, or operation) sends a prompt; the LLM may return both
  a message and a set of tool calls; Orpius executes the permitted tools
  and feeds the results back to the model until it stops requesting
  tools. Tools are the mechanism by which an agent can act on its
  environment. For cross-node communication, Orpius provides the
  \emph{OperationRelay} as a built-in tool. When enabled for an agent,
  OperationRelay allows that agent to call a remote Operation,
  potentially on a different Orpius server, using an external identifier
  and access key. The relay preserves conversation context, normalises
  results into the standard tool-calling protocol, and enforces least
  privilege: a remote Operation can be invoked only if its URL and
  credentials have been explicitly provided.
\end{enumerate}

\subsubsection{Local tool-calling loop}\label{local-tool-calling-loop}

\pandocbounded{\includegraphics[keepaspectratio,alt={}]{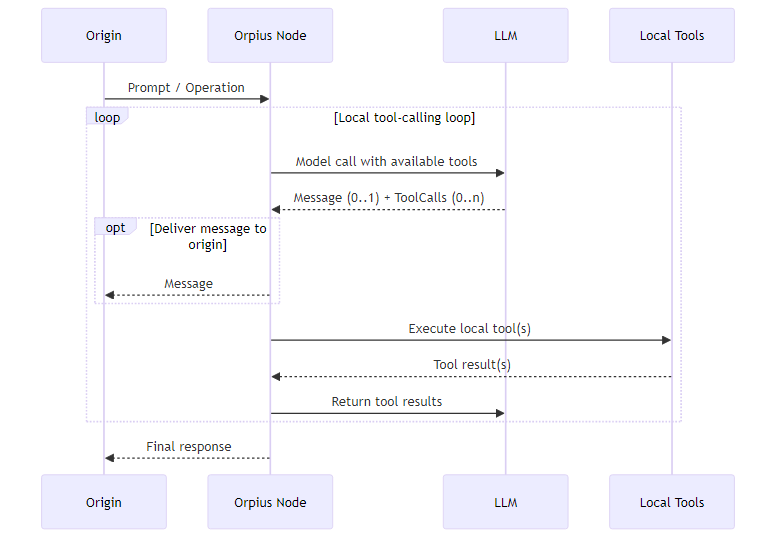}}

\emph{Figure 3: Local tool-calling loop within a single Orpius node. The
LLM may emit messages and tool calls; Orpius executes permitted tools
and returns results until the model stops requesting further actions.}

\subsubsection{Cross-node coordination with
OperationRelay}\label{cross-node-coordination-with-operationrelay}

\pandocbounded{\includegraphics[keepaspectratio,alt={}]{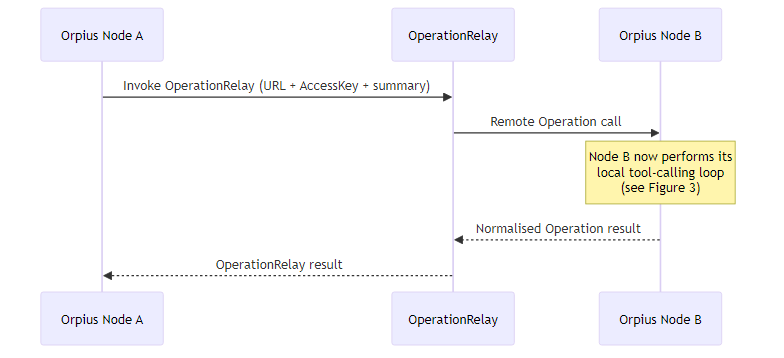}}

\emph{Figure 4: OperationRelay enabling cross-node agent interaction. A
remote Operation runs its own LLM--tool loop and returns a normalised
result to the originating node, preserving context and tool semantics.}

Taken together, these properties mean that:

\begin{itemize}
\tightlist
\item
  each organisation retains full control over its node, including which
  agents exist, which tools they may use, and which Operations are
  exposed externally;
\item
  \emph{data locality} is enforced at the platform level, as agents can
  only access files and tools within their own node unless an explicit
  cross-node Operation is configured; and
\item
  \emph{federated reasoning} emerges from the composition of local
  agents, tools, and OperationRelay calls, rather than from any shared
  global database or schema.
\end{itemize}

The remainder of the paper instantiates these capabilities in a concrete
three-node scenario, showing how the node model, tool orchestration, and
OperationRelay enable multi-agent coordination across organisational
boundaries using only natural-language messages and pseudonymous tokens.
The resulting workflow is decentralised and privacy-preserving, with
each organisation contributing domain-specific reasoning while
maintaining strict data separation.

\section{Prototype Architecture}\label{prototype-architecture}

\subsection{High-Level Structure}\label{high-level-structure}

The architecture includes:

\begin{itemize}
\item
  \textbf{Clinic Agent}

  \begin{itemize}
  \tightlist
  \item
    Generates pseudonymous tokens
  \item
    Summarises clinical observations
  \item
    Sends natural-language coverage inquiries to the Insurer
  \end{itemize}
\item
  \textbf{Insurer Agent}

  \begin{itemize}
  \tightlist
  \item
    Matches patient token to local enrolment
  \item
    Checks coverage rules
  \item
    When needed, calls the Specialist agent for clinical appropriateness
  \item
    Returns a combined natural-language verdict
  \end{itemize}
\item
  \textbf{Specialist Agent}

  \begin{itemize}
  \tightlist
  \item
    Consults local guidance
  \item
    Provides appropriateness recommendations
  \item
    Never receives identifiers or tokens
  \end{itemize}
\end{itemize}

This forms a \emph{three-node distributed agent chain}, orchestrated
exclusively through natural-language descriptions of a case.

\pandocbounded{\includegraphics[keepaspectratio,alt={}]{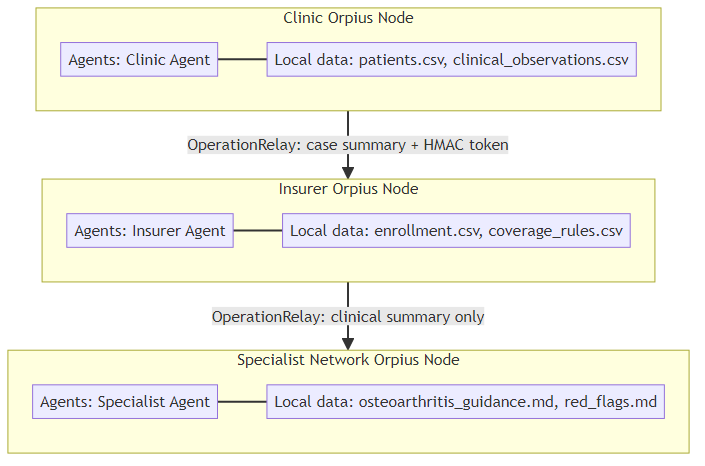}}

\emph{Figure 5: Architecture of the distributed Orpius prototype with
three independent nodes linked by OperationRelay.}

\subsection{Local Data and Boundaries}\label{local-data-and-boundaries}

Each organisation holds only the data relevant to its role:

\begin{itemize}
\item
  \textbf{Clinic} Synthetic CSV files describing symptoms, severity,
  duration, and prior management.
\item
  \textbf{Insurer} Synthetic enrolment table and coverage rules.
\item
  \textbf{Specialist Network} Extracts from synthetic osteoarthritis
  guidelines and red-flag criteria.
\end{itemize}

No system has direct access to another's storage. Every agent relies
solely on:

\begin{itemize}
\tightlist
\item
  the incoming natural-language summary,
\item
  local files,
\item
  and any follow-up responses via OperationRelay
\end{itemize}

This arrangement intentionally mirrors real inter-organisational
constraints.

\section{Agents and Responsibilities}\label{agents-and-responsibilities}

This section summarises the refined behaviour of the three agents as
implemented in the prototype, using the actual operation instructions
provided in the appendices.

\subsection{Clinic Agent}\label{clinic-agent}

The Clinic Agent performs the following steps:

\begin{enumerate}
\def\labelenumi{\arabic{enumi}.}
\item
  \textbf{Gather case facts from local CSV files.} It loads symptom
  severity, functional limitations, duration, and prior conservative
  management.
\item
  \textbf{Determine the proposed intervention.} If the external request
  specifies a treatment, that is used; otherwise, it selects the most
  plausible intervention based on the data.
\item
  \textbf{Generate a pseudonymous patient token.} The agent computes

\begin{Shaded}
\begin{Highlighting}[]
\NormalTok{patientToken = HMACSHA256(key, UPPER(patientId))}
\end{Highlighting}
\end{Shaded}

  and never transmits the actual patient ID.
\item
  \textbf{Send a coverage inquiry in natural language} to the Insurer
  Agent using OperationRelay. The message includes only:

  \begin{itemize}
  \tightlist
  \item
    patient token
  \item
    symptoms
  \item
    functional limitations
  \item
    prior management
  \item
    proposed treatment
  \end{itemize}
\item
  \textbf{Return the insurer's response to the user.}
\end{enumerate}

The Clinic Agent never discloses identity or raw CSV paths and performs
all work in a single step without clarifying questions.

\subsection{Insurer Agent}\label{insurer-agent}

The Insurer Agent receives natural-language coverage inquiries
containing:

\begin{itemize}
\tightlist
\item
  a patient token
\item
  a clinical summary
\item
  the proposed treatment
\end{itemize}

Using its local data:

\begin{enumerate}
\def\labelenumi{\arabic{enumi}.}
\item
  Match the patient token in \texttt{enrollment.csv}.
\item
  Determine coverage rules for the proposed treatment.
\item
  If coverage depends on clinical appropriateness, call the Specialist
  Agent via OperationRelay. Crucially, the insurer does not forward the
  patient token to the specialist, only a clinical narrative summary.
\item
  Combine coverage and appropriateness into a single, concise
  natural-language response.
\end{enumerate}

The insurer never reconstructs identity or stores the patient token
beyond the immediate evaluation.

\subsection{Specialist Agent}\label{specialist-agent}

The Specialist Agent receives a narrative description of:

\begin{itemize}
\tightlist
\item
  condition and severity
\item
  duration
\item
  functional limitations
\item
  prior conservative management
\item
  proposed treatment
\end{itemize}

It consults its local guideline extracts and returns:

\begin{itemize}
\tightlist
\item
  a recommendation (appropriate now / appropriate after steps / not
  currently appropriate)
\item
  a short justification
\item
  optional next steps
\end{itemize}

The Specialist Agent requires no identifiers, tokens, or conversation
handles and makes no external calls unless explicitly permitted.

\section{Privacy and Tokenisation
Strategy}\label{privacy-and-tokenisation-strategy}

\subsection{Principles}\label{principles}

The design follows three principles:

\begin{enumerate}
\def\labelenumi{\arabic{enumi}.}
\item
  \textbf{Data Minimisation} Each organisation exchanges only what is
  necessary for its task.
\item
  \textbf{Locality} All detailed records remain inside the originating
  Orpius deployment.
\item
  \textbf{Pseudonymous Linking} Patient identity is replaced with a
  one-way HMAC token, enabling linkage by the insurer while preventing
  re-identification by other agents.
\end{enumerate}

\subsection{Token Computation}\label{token-computation}

The Clinic computes:

\begin{verbatim}
patientToken = HMACSHA256(secret_key, UPPER(TRIM(patientId)))
\end{verbatim}

\begin{itemize}
\tightlist
\item
  Secret key stored as a managed Orpius secret
\item
  Token formatted as lowercase hex
\item
  Identity never transmitted externally
\end{itemize}

This allows the Insurer to link cases deterministically \emph{only} for
the purpose of evaluating coverage.

Other linkage strategies (for example multi-step handshakes or
insurer-issued ephemeral handles) are possible. However, all such
designs still require the Clinic to supply a cross-node reference that
the Insurer can resolve. In this prototype we adopt the deterministic
one-way token because it provides privacy-preserving linkage without
additional round-trips, shared state, or protocol complexity, and it is
sufficient to illustrate the architectural patterns that are the focus
of this work.

\subsection{Natural-Language Message
Passing}\label{natural-language-message-passing}

Instead of structured schemas, messages are exchanged in concise
natural-language paragraphs containing:

\begin{itemize}
\tightlist
\item
  symptom class (e.g., mild/moderate/severe)
\item
  functional limitations
\item
  approximate duration
\item
  prior conservative treatment
\item
  proposed intervention
\end{itemize}

This approach reduces coupling and allows each agent to reason with
local domain knowledge without enforcing a rigid data model.

\section{Formalising Data Locality and Cross-Node Information
Flow}\label{formalising-data-locality-and-cross-node-information-flow}

To clarify the architectural guarantees of the prototype, we can express
the data-locality and message-passing constraints using a simple
mathematical model. Let each participating organisation be represented
as a node:

\[
\mathcal{N} = \{\text{Clinic},\, \text{Insurer},\, \text{Specialist}\}.
\]

Each node \(n \in \mathcal{N}\) holds a private dataset \(D_n\). These
datasets are disjoint:

\[
D_i \cap D_j = \varnothing
\qquad\text{for all } i \neq j.
\]

No dataset is ever transmitted or partially revealed to another
organisation. Instead, each node produces a natural-language summary
based solely on its local data and the incoming request. We model this
as a local summarisation function:

\[
m_{i \to j} = f_i(D_i,\, q),
\]

where \(q\) is the inbound natural-language request and \(m_{i \to j}\)
is the outbound message delivered to node \(j\). The function \(f_i\)
represents the LLM-guided reasoning and tool-assisted lookups performed
within node \(i\). Importantly, \(m_{i \to j}\) is constrained to
contain only high-level clinical or administrative descriptors and, in
the case of Clinic to Insurer communication, a pseudonymous token rather
than any raw identifier.

The data-locality requirement can therefore be expressed as:

\[
D_i \cap m_{i\to j} = \varnothing
\qquad\text{and}\qquad
m_{i\to j}\big|_{\text{identity}} = \varnothing.
\]

Cross-organisation coordination is captured as a directed exchange of
summaries:

\[
\text{Clinic}
  \xrightarrow{m_{C \to I}}
\text{Insurer}
\longrightarrow
\text{Specialist}
  \xrightarrow{m_{S \to I}}
\text{Insurer}
\longrightarrow
\text{Clinic}.
\]

All reasoning is local:

\[
r_i = R_i(D_i,\, m_{\ast \to i}),
\]

where \(R_i\) denotes the local decision process combining LLM
inferences and tool execution inside node \(i\). No node ever
reconstructs or infers another node's dataset, since every message is a
lossy projection of local information.

Taken together, these relations define a system in which cooperation
arises entirely from the composition of summary-producing functions
\(\{f_i\}\) rather than from any shared global state. This provides a
formal characterisation of the prototype's strict data-locality and
privacy-preserving message-passing design.

\section{Example Interaction Flow}\label{example-interaction-flow}

Below is the simplified trace of a real prototype run, using synthetic
data. The specific phrasing will vary depending on the model.

\subsubsection{Step 1: User asks the Clinic
Agent}\label{step-1-user-asks-the-clinic-agent}

The user triggers an Orpius operation, e.g.:

\begin{verbatim}
"Confirm coverage for CLN-0001"
\end{verbatim}

\subsubsection{Step 2: Clinic Agent Constructs
Summary}\label{step-2-clinic-agent-constructs-summary}

The Clinic Agent loads the clinical data and produces a natural-language
request:

\begin{verbatim}
Coverage inquiry for patient_token=<computed token>
Presentation: moderate knee pain limiting stairs 
              and prolonged standing; NSAID 2 weeks; home exercises.
Proposed treatment: conservative management.
Please advise coverage and any prerequisites.
\end{verbatim}

\subsubsection{Step 3: Insurer Agent Evaluates
Coverage}\label{step-3-insurer-agent-evaluates-coverage}

The Insurer checks enrolment, coverage rules, and, if necessary, queries
the Specialist Agent.

\subsubsection{Step 4: Specialist Agent
Responds}\label{step-4-specialist-agent-responds}

A typical response:

\begin{verbatim}
Recommendation: Appropriate now.
Reasoning: Standard conservative management is consistent with moderate symptoms
and early functional limitation.
Next steps: Continue structured physiotherapy if not already completed.
(ref: osteoarthritis_knee_guidance.md)
\end{verbatim}

\subsubsection{Step 5: Insurer Agent Returns Final
Verdict}\label{step-5-insurer-agent-returns-final-verdict}

Example outcome:

\begin{verbatim}
Coverage: Not covered
Clinical Appropriateness: Appropriate now
Summary: Conservative management is clinically appropriate but is not covered
         under the current plan.
Next Steps: Consider reviewing plan options or alternative interventions.
\end{verbatim}

\pandocbounded{\includegraphics[keepaspectratio]{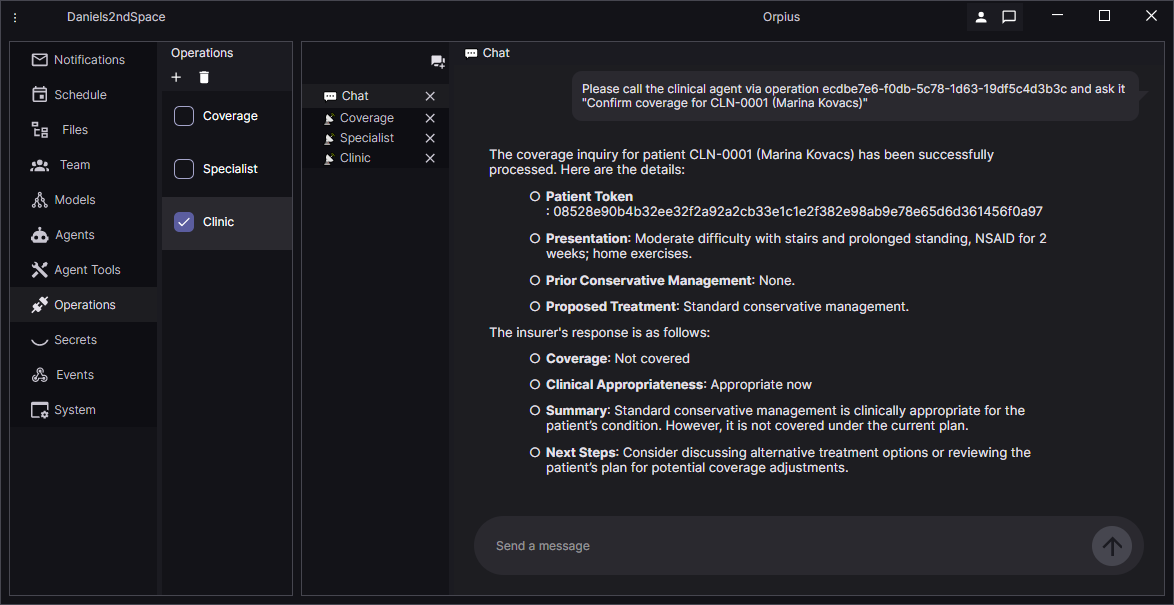}}

\emph{Figure 6: Example of a user request routed through the Clinic
Agent, with downstream calls to Insurer and Specialist Agents.}

\section{Limitations}\label{limitations}

This work is intentionally narrow in scope. Several limitations reflect
the prototype's exploratory nature:

\subsubsection{No Clinical Validation}\label{no-clinical-validation}

The system is not clinically evaluated. The guideline extracts,
synthetic data, and recommendations are illustrative only.

\subsubsection{No Formal Evaluation or
Benchmarking}\label{no-formal-evaluation-or-benchmarking}

We did not measure:

\begin{itemize}
\tightlist
\item
  accuracy
\item
  latency
\item
  cost
\item
  robustness
\item
  failure handling
\end{itemize}

Only a small number of manual test runs were conducted to verify
functional behaviour.

\subsubsection{Small Synthetic Dataset}\label{small-synthetic-dataset}

All CSV files provided in the appendices are minimal and simplified.
Real clinical, insurance, and specialist datasets are vastly more
complex.

\subsubsection{Natural-Language
Variability}\label{natural-language-variability}

Because agents interact through free-form natural language, the system
behaviour depends on LLM phrasing. We did not evaluate stability across
models, temperature settings, or prompt variations.

\subsubsection{Limited Cross-Organisation
Complexity}\label{limited-cross-organisation-complexity}

The prototype shows only one type of cross-system coordination (coverage
clarification). More complex workflows might reveal new constraints or
unintended behaviours.

\subsubsection{No Adversarial or Security
Testing}\label{no-adversarial-or-security-testing}

The prototype does not evaluate security aspects beyond enforcing
data-locality boundaries. We did not assess resilience to adversarial
behaviour, including prompt-injection attacks, malicious summarisation,
identifier leakage, or cross-system inference risks. A comprehensive
security analysis lies outside the scope of this exploratory study.

\section{Future Work}\label{future-work}

Future directions include:

\subsubsection{Expanded Clinical
Scenarios}\label{expanded-clinical-scenarios}

Supporting multiple conditions, richer reasoning chains, and additional
specialist domains.

\subsubsection{Robust Evaluation
Framework}\label{robust-evaluation-framework}

Designing controlled experiments to measure:

\begin{itemize}
\tightlist
\item
  correctness
\item
  stability
\item
  inter-agent consistency
\item
  token privacy preservation
\item
  performance across multiple model providers
\end{itemize}

\subsubsection{Formal Security Analysis}\label{formal-security-analysis}

A full assessment of adversarial robustness is needed. This includes
evaluating exposure to prompt-injection vectors, malicious
summarisation, identifier leakage, cross-system inference attacks, and
the security guarantees of OperationRelay in distributed deployments.

\subsubsection{Schema-Light or Schema-Aware
Mediation}\label{schema-light-or-schema-aware-mediation}

Maintaining natural-language flexibility while introducing optional
structure (e.g., semantic tagging or lightweight reporting templates).

\subsubsection{Enhanced Privacy
Guarantees}\label{enhanced-privacy-guarantees}

Exploration of:

\begin{itemize}
\tightlist
\item
  differential privacy for summarisation
\item
  formal guarantees around pseudonymous identifiers
\item
  multi-key token derivation for multi-insurer settings
\end{itemize}

\subsubsection{Distributed Orpius Mesh}\label{distributed-orpius-mesh}

Orpius already supports isolated deployments; future work may include
higher-level primitives for secure federation, multi-hop workflows, and
peer discovery.

\subsubsection{Human-in-the-Loop
Oversight}\label{human-in-the-loop-oversight}

Supporting checkpoints where human reviewers can audit decisions or
override automated inferences.

\section{Conclusion}\label{conclusion}

We presented a proof-of-concept demonstration of natural-language
multi-agent coordination across distributed Orpius systems, modelling
inter-organisational collaboration between a Clinic, Insurer, and
Specialist Network.

The system uses:

\begin{itemize}
\tightlist
\item
  Strict data locality
\item
  Lightweight pseudonymisation to avoid sharing identifiers
\item
  Orpius OperationRelay for cross-system messaging
\item
  Local reasoning against isolated datasets
\item
  LLM-driven agents performing summarisation and decision support
\end{itemize}

This work does not aim to deliver clinical correctness or
production-ready capability. Rather, it shows that decentralised,
privacy-preserving, natural-language coordination between autonomous
agents is feasible using Orpius's architectural model. This prototype
illustrates a possible foundation for inter-organisational agent systems
that operate without data sharing, suggesting a direction for future
architectures in regulated domains.

We hope this demonstration encourages further exploration of distributed
agent systems, privacy-preserving communication patterns, and real-world
integration scenarios.

\section{Appendices}\label{appendices}

Below are the abbreviated appendices containing only the synthetic data
and operational instructions used in the prototype.

\begin{center}\rule{0.5\linewidth}{0.5pt}\end{center}

\subsection{Appendix A: Synthetic Data
Samples}\label{appendix-a-synthetic-data-samples}

\subsubsection{A.1 Clinic Data}\label{a.1-clinic-data}

\textbf{clinical\_observations.csv}

\begin{verbatim}
patient_id,symptom_class,duration_weeks,functional_limitation,prior_conservative_tx
CLN-0001,moderate,12,difficulty stairs and prolonged standing,NSAID_2_weeks;home_exercises
CLN-0002,mild,6,pain after long walks,acetaminophen_2_weeks
CLN-0003,severe,24,pain at rest; limited ROM,NSAID_3_weeks;physio_4_weeks
CLN-0004,moderate,10,difficulty rising from chair,NSAID_2_weeks
CLN-0005,moderate,14,stairs; standing >20min,NSAID_2_weeks;physio_6_weeks
\end{verbatim}

\textbf{patients.csv}

\begin{verbatim}
patient_id,full_name,dob,notes
CLN-0001,Marina Kovacs,1968-08-12,"Right knee pain, gradual onset"
CLN-0002,John Armitage,1975-04-03,Knee discomfort after long walks
CLN-0003,Sofia Patel,1982-11-19,Chronic knee pain with reduced ROM
CLN-0004,Luca Meier,1961-02-07,Difficulty rising from chair
CLN-0005,Elena Rossi,1990-06-28,Pain with stairs and prolonged standing
\end{verbatim}

\subsubsection{A.2 Insurer Data}\label{a.2-insurer-data}

\textbf{coverage\_rules.csv}

\begin{verbatim}
plan_id,treatment_code,covered,prerequisites
PLAN-A,knee_hyaluronic_injection,yes,physiotherapy_6_weeks;failed_simple_analgesia
PLAN-B,knee_hyaluronic_injection,yes,failed_simple_analgesia
PLAN-C,knee_hyaluronic_injection,no,—
PLAN-A,physiotherapy_course,yes,—
PLAN-B,physiotherapy_course,yes,—
PLAN-C,physiotherapy_course,yes,—
\end{verbatim}

\textbf{enrollment.csv}

\begin{verbatim}
subject_token,insurance_number,plan_id,status
08528e90b4b32ee32f2a92a2cb33e1c1e2f382e98ab9e78e65d6d361456f0a97,INS-441122,PLAN-A,active
8e5be9b3d9b4b25894e16137b1038c99dde5fcc2c7fe4ef75d53b69696655f3d,INS-771102,PLAN-B,active
470b4bc40a6314f68355838ef062527985b57ea6b4bce3930b12adbc0b0d7e65,INS-555901,PLAN-C,active
80f2912672068583432994fa4a75ce3fb9d1adaa02df386d622369c381e48763,INS-880014,PLAN-A,active
966a4a55035f496d38d1dab53af49b549f4c9a921989c5799244b0db94614aed,INS-993377,PLAN-B,active
\end{verbatim}

\subsubsection{A.3 Specialist Guidance
Extracts}\label{a.3-specialist-guidance-extracts}

\textbf{osteoarthritis\_knee\_guidance.md}

\begin{verbatim}
# Knee Osteoarthritis – Practical Guidance

## Symptom Severity Bands
- Mild: intermittent pain, minimal function impact
- Moderate: persistent pain >= 6–8 weeks, functional limitation
- Severe: pain at rest, substantial functional loss

## Conservative Management Ladder
1. Education, activity modification, weight optimisation
2. Simple analgesia 2–3 weeks
3. Physiotherapy 6 weeks
4. Consider intra-articular options if function remains impaired

## Intra-Articular Hyaluronic Acid (HA) Injection
- Consider after steps 1–3 when pain/function remain limiting

## Notes
- Document adherence to home programmes
- Functional limitation weighs toward escalation
\end{verbatim}

\textbf{osteoarthritis\_knee\_red\_flags.md}

\begin{verbatim}
# Red Flags – Knee Pain
- Acute trauma with inability to bear weight
- Suspected infection
- Locked knee
- Suspected fracture or major ligament injury
\end{verbatim}

\begin{center}\rule{0.5\linewidth}{0.5pt}\end{center}

\subsection{Appendix B: Operation Instructions
(Abbreviated)}\label{appendix-b-operation-instructions-abbreviated}

These are lightly condensed versions of the operational instructions the
agents used.

\subsubsection{B.1 Clinic Agent
Instructions}\label{b.1-clinic-agent-instructions}

\begin{itemize}
\item
  Never transmit identity; compute \texttt{patientToken} using HMAC.
\item
  Load local CSV data to construct a clinical summary.
\item
  Choose or infer the proposed treatment.
\item
  Send a natural-language coverage inquiry via OperationRelay:

  \begin{itemize}
  \tightlist
  \item
    include token, symptoms, duration, functional limitations, prior
    management
  \item
    exclude identity
  \end{itemize}
\item
  Return insurer response to user.
\end{itemize}

\subsubsection{B.2 Insurer Agent
Instructions}\label{b.2-insurer-agent-instructions}

\begin{itemize}
\tightlist
\item
  Match token in \texttt{enrollment.csv}.
\item
  Evaluate coverage rules.
\item
  If appropriateness unclear, call Specialist Agent with a clinical
  summary (no token).
\item
  Combine coverage + appropriateness into one concise message.
\item
  Maintain strict privacy constraints.
\end{itemize}

\subsubsection{B.3 Specialist Agent
Instructions}\label{b.3-specialist-agent-instructions}

\begin{itemize}
\item
  Receive only a clinical summary.
\item
  Consult local guideline extracts.
\item
  Respond with:

  \begin{itemize}
  \tightlist
  \item
    Recommendation (appropriate now / after steps / not currently
    appropriate)
  \item
    Reasoning (1--3 sentences)
  \item
    Optional next steps
  \end{itemize}
\item
  Never receive or use identifiers.
\end{itemize}

\begin{center}\rule{0.5\linewidth}{0.5pt}\end{center}

\section*{References}\label{bibliography}
\addcontentsline{toc}{section}{References}

\protect\phantomsection\label{refs}

\end{document}